\documentclass[11pt, a4paper]{article}

\usepackage[utf8]{inputenc}
\usepackage[T1]{fontenc}
\usepackage{amsmath, amssymb, amsthm}
\usepackage{graphicx}
\usepackage{booktabs}
\usepackage{algorithm}
\usepackage{algorithmic}
\usepackage{hyperref}
\usepackage{xcolor}
\usepackage{caption}
\usepackage{subcaption}
\usepackage[affil-it]{authblk} 
\usepackage{float}
\usepackage{tikz}
\usetikzlibrary{shapes, arrows.meta, positioning, calc}
\usepackage[margin=1in]{geometry} 
\hypersetup{
    colorlinks=true,
    linkcolor=blue,
    filecolor=magenta,      
    urlcolor=cyan,
    citecolor=red,
}

\title{\textbf{Conditional Morphogenesis: Emergent Generation of Structural Digits via Neural Cellular Automata}}
\author{Ali Sakour}
\affil{Department of Computer and Automatic Control Engineering, \\ Faculty of Mechanical and Electrical Engineering, \\ Lattakia University, Lattakia, Syria \\ \texttt{alisakour78@gmail.com}}

\date{\today}

\begin{document}

\maketitle

\begin{abstract}
\noindent Biological systems exhibit remarkable morphogenetic plasticity, where a single genome can encode various specialized cellular structures triggered by local chemical signals. In the domain of Deep Learning, Differentiable Neural Cellular Automata (NCA) have emerged as a paradigm to mimic this self-organization. However, existing NCA research has predominantly focused on continuous texture synthesis or single-target object recovery, leaving the challenge of \textit{class-conditional structural generation} largely unexplored. 

In this work, we propose a novel \textbf{Conditional Neural Cellular Automata (c-NCA)} architecture capable of growing distinct topological structures—specifically MNIST digits—from a single generic seed, guided solely by a spatially broadcasted class vector. Unlike traditional generative models (e.g., GANs, VAEs) that rely on global reception fields, our model enforces strict locality and translation equivariance. We demonstrate that by injecting a one-hot condition into the cellular perception field, a single set of local rules can learn to break symmetry and self-assemble into ten distinct geometric attractors. Experimental results show that our c-NCA achieves stable convergence, correctly forming digit topologies from a single pixel, and exhibits robustness characteristic of biological systems. This work bridges the gap between texture-based NCAs and structural pattern formation, offering a lightweight, biologically plausible alternative for conditional generation.

\vspace{0.5em}
\noindent \textbf{Code Availability:} The source code and pretrained models are available at: \url{https://github.com/alisakour/Conditional-NCA-Digits}

\vspace{0.5em}
\end{abstract}

\section{Introduction}

In biological systems, complex organisms emerge from a single cell through a process known as \textit{morphogenesis}. Unlike man-made machinery, which is assembled by external agents following a global blueprint, biological structures self-organize through local interactions driven by a shared genetic code \cite{mordvintsev2020growing}. Every cell contains the same instructions, yet they differentiate into distinct tissues—bone, muscle, or nerve—based on local chemical signaling and environmental context. 

In the realm of Deep Learning, Generative Adversarial Networks (GANs) and Variational Autoencoders (VAEs) have achieved remarkable success in image synthesis. However, these models operate under a "top-down" paradigm, utilizing global receptive fields where a central controller dictates pixel values. While effective, this lacks the robustness, decentralized repair capabilities, and biological plausibility of natural systems.

Differentiable Neural Cellular Automata (NCA) have recently bridged this gap, demonstrating that deep learning can optimize the local update rules of a cellular grid to grow target patterns, such as emojis or textures. Despite these advancements, a significant limitation remains in the literature: \textbf{Specificity}. Existing NCA models are typically "specialists"—trained to grow a single static shape or an infinite texture. The challenge of creating a "generalist" NCA—one that can morph into multiple distinct geometric structures based on a conditional signal—remains largely unsolved for structural data.

In this paper, we address this limitation by proposing a \textbf{Conditional Neural Cellular Automata (c-NCA)}. We investigate whether a single set of local parameters can encode the topological "DNA" for ten distinct handwritten digits (MNIST) and selectively express them via a spatially broadcasted control vector. Unlike texture synthesis, digit generation requires strict adherence to spatial constraints (e.g., the precise curvature of a '2' versus the loop of a '6'). Our work provides a proof-of-concept that local, translation-equivariant rules are sufficient for class-conditional structural morphogenesis, opening new avenues for resilient, biologically-inspired generative AI.

\section{Related Work}

\subsection{Neural Cellular Automata (NCA)}
The concept of Cellular Automata (CA) dates back to von Neumann and Conway's "Game of Life," demonstrating how complex global behaviors emerge from simple local rules. Recently, Mordvintsev et al. \cite{mordvintsev2020growing} introduced \textit{Differentiable Neural Cellular Automata}, enabling the learning of local update rules via backpropagation. Their work successfully demonstrated the ability to grow specific target images (e.g., emojis) from a single seed and repair them after damage. However, the original NCA formulation is typically trained to produce a \textit{single} static attractor or a continuous texture pattern \cite{niklasson2021self}. It lacks an intrinsic mechanism for class-conditional generation, meaning a separate model must be trained for each target shape. Our work extends this framework by introducing a conditional perception mechanism, allowing a single set of parameters to generate multiple distinct topologies (MNIST digits) based on an external control vector.

\subsection{Conditional Generative Models}
In the domain of image generation, Conditional Generative Adversarial Networks (cGANs) \cite{mirza2014conditional} and Conditional Variational Autoencoders (cVAEs) \cite{sohn2015learning} are the standard benchmarks. These models condition the generation process on class labels (e.g., via one-hot encoding) to produce specific digits. However, these architectures rely on global processing (e.g., fully connected layers or standard CNNs with global receptive fields) to map latent vectors to pixels directly or hierarchically. In contrast, our approach treats generation as a \textit{biological morphogenesis process}. There is no global "generator" that sees the whole image; instead, every cell acts independently based on local chemical signals and the broadcasted condition, offering a radically different, biologically plausible perspective on generative modeling.

\subsection{Structural Morphogenesis vs. Texture Synthesis}
While several extensions of NCA have explored 3D voxel growth or texture synthesis, the application of NCAs to strictly constrained 2D structural data like handwritten digits remains underexplored. Unlike textures, which are translation-invariant and repetitive, digits require precise spatial topology (e.g., the loop of a '9' vs. the stroke of a '1'). To the best of our knowledge, this is one of the first works to successfully apply a Conditional NCA to the full MNIST dataset, demonstrating that strictly local interactions, when conditioned properly, can satisfy global structural constraints of multiple distinct classes.
\subsection{Conditional Control Mechanisms in NCA}
The concept of conditional NCA, where the growth process is guided by an external signal, has been explored previously. Notably, Frans \cite{frans2021stampca} introduced StampCA, where conditioning information is encoded exclusively into the state of the initial seed cell at $t=0$. The cellular automata must then learn to propagate this initial information across the grid.

Our work differs and extends this approach in two fundamental aspects. First, our conditioning mechanism provides \textbf{persistent guidance}; the one-hot class vector is concatenated with the perception field at every time step, acting as a continuous environmental signal rather than a one-time instruction. Second, and more critically, we rigorously demonstrate the emergent property of \textbf{universal self-repair}. We show that our c-NCA can recover from severe stochastic degradation across all ten digit classes, a property not investigated in the StampCA framework. This highlights the robustness of the learned attractors in our model.

\subsection{Advanced Conditional and Adaptive NCAs}
Recent research has extended the NCA paradigm towards more complex forms of control and adaptation. 

Palm et al. \cite{palm2022variational} proposed \textbf{Variational Neural Cellular Automata (VNCA)}, which combines NCAs with Variational Autoencoders (VAEs) to learn a smooth latent space of NCA parameters. Conditioning is achieved by sampling a latent vector $\mathbf{z}$ and using a hypernetwork to generate the NCA weights. While this enables impressive interpolation between shapes, it is computationally intensive and differs from our approach, where we utilize a single, fixed set of rules and inject the condition directly into the cell's perception. Our method is more aligned with biological systems where the "laws of physics" (the rules) are constant, but "environmental signals" (the condition) vary.

Another line of work focuses on adapting the perception mechanism itself. Grattarola et al. \cite{grattarola2022adanca} introduced \textbf{AdaNCA}, where the perception kernel is dynamic, allowing cells to change their receptive field size during growth. This solves the challenge of long-range information propagation. Our work is orthogonal and complementary; while AdaNCA adapts the "how" of perception, our c-NCA controls the "what" of morphogenesis based on an external goal.

Closest to our work in spirit is the \textbf{Goal-Guided NCA} by Sudhakaran et al. \cite{sudhakaran2022goal}, where a "goal map" is persistently provided to the NCA to guide its behavior in reinforcement learning tasks. This is similar to our persistent conditioning mechanism. However, their focus is on achieving functional goals (e.g., navigating a maze), whereas our work is centered on achieving high-fidelity structural and topological targets (i.e., precise digit shapes). Critically, our exhaustive demonstration of self-repair across all ten classes provides a rigorous validation of structural robustness not explored in these goal-seeking contexts.
% --- 3. Methodology ---
\section{Methodology}

We formulate the problem of conditional digit generation as learning the local update rules of a discrete-time dynamical system. The system operates on a regular grid lattice, where the global evolution of the state emerges solely from local interactions.

\subsection{State Space Representation}
Let the state of the grid at time $t$ be denoted by $S_t \in \mathbb{R}^{H \times W \times C}$, where $H=W=28$ are the spatial dimensions and $C=16$ is the channel depth. The channels are partitioned into two functional groups:
\begin{equation}
    S_t^{(x,y)} = [\underbrace{r, g, b, \alpha}_{\text{Visible State}}, \underbrace{h_1, \dots, h_{12}}_{\text{Hidden State}}]^T
\end{equation}
The first four channels represent the visible RGBA values, while the remaining 12 channels act as a latent "chemical" memory, facilitating long-range communication through local propagation. The grid is initialized as $S_0 = \mathbf{0}$, except for a single "seed" activation at the center $S_0^{(H/2, W/2)} = [0, 0, 0, 1, \dots, 0]^T$.

\subsection{Learnable Perception}
Unlike traditional Cellular Automata which use fixed interaction rules, our model learns to perceive its neighborhood via a convolution operation. To maintain translation equivariance—a critical property for biological plausibility—we employ a learnable $3 \times 3$ depthwise convolution. For a cell at position $u$, the perception vector $z_u$ is given by:
\begin{equation}
    z_u = K_{perc} * S_t(u) \quad \in \mathbb{R}^{3C}
\end{equation}
where $*$ denotes the convolution operator and $K_{perc}$ represents the learned filters. This operation allows each cell to sense the state gradients and intensity of its immediate Moore neighborhood.

\subsection{Conditional Control Mechanism}
To enable class-specific morphogenesis, we introduce a conditioning vector $\mathbf{c} \in \{0,1\}^{10}$, representing the one-hot encoded target digit. Since the update rule is local, the global condition must be made available to every cell. We achieve this by spatially broadcasting $\mathbf{c}$ to match the grid dimensions, resulting in a condition map $\mathcal{C} \in \mathbb{R}^{H \times W \times 10}$.

The input to the update policy is the concatenation of the local perception and the global condition:
\begin{equation}
    \mathbf{x}_{in} = \text{Concat}(z_u, \mathbf{c})
\end{equation}

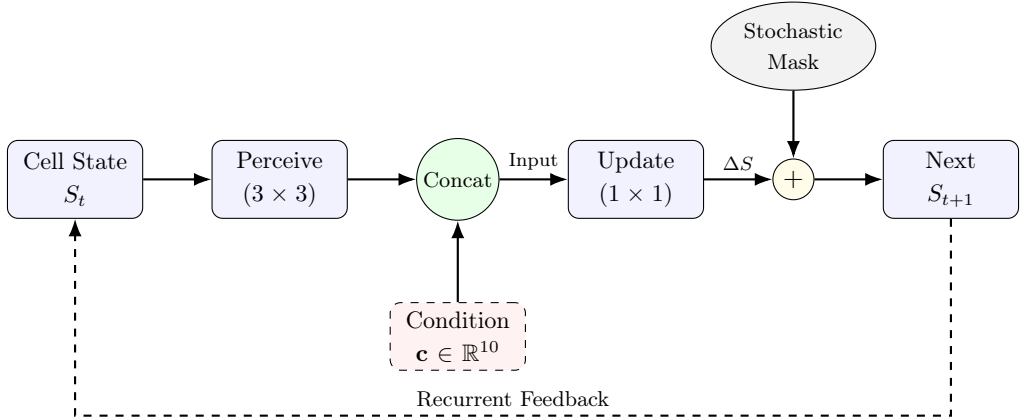
\begin{figure}[ht]
    \centering
    \begin{tikzpicture}[
        scale=0.9, transform shape,
        node distance=1.2cm and 1.2cm, 
        auto,
        block/.style={
            rectangle, 
            draw, 
            fill=blue!5, 
            text width=4.5em, 
            text centered, 
            rounded corners, 
            minimum height=3em,
            font=\small
        },
        cond/.style={
            rectangle, 
            draw, 
            fill=red!5, 
            text width=4.5em, 
            text centered, 
            rounded corners, 
            minimum height=2.5em,
            dashed,
            font=\small
        },
        line/.style={
            draw, 
            -Latex,
            thick
        },
        sum/.style={
            circle, 
            draw, 
            fill=yellow!10, 
            inner sep=0pt, 
            minimum size=6mm
        }
    ]

    \node [block] (state) {Cell State \\ $S_t$};
    \node [block, right=1cm of state] (perc) {Perceive \\ ($3\times3$)};
    
    % Concat node
    \node [circle, draw, fill=green!10, right=1cm of perc, inner sep=2pt, font=\footnotesize] (concat) {Concat};
    
    \node [block, right=1cm of concat] (mlp) {Update \\ ($1\times1$)};
    \node [sum, right=1cm of mlp] (plus) {$+$};
    \node [block, right=1cm of plus] (next) {Next \\ $S_{t+1}$};

    \node [cond, below=1.2cm of concat] (condition) {Condition \\ $\mathbf{c} \in \mathbb{R}^{10}$};

    \node [draw, ellipse, fill=gray!10, above=1cm of plus, font=\footnotesize, align=center] (stoch) {Stochastic \\ Mask};

    \path [line] (state) -- (perc);
    \path [line] (perc) -- (concat);
    \path [line] (concat) -- node [font=\scriptsize, above] {Input} (mlp);
    \path [line] (mlp) -- node [font=\scriptsize, above] {$\Delta S$} (plus);
    \path [line] (plus) -- (next);

    \path [line] (condition) -- (concat);
    \path [line] (stoch) -- (plus);

    \draw [line, dashed] (next.south) -- ++(0,-2.9) -| (state.south) 
        node [pos=0.25, above, font=\footnotesize] {Recurrent Feedback};

    \end{tikzpicture}
    \caption{Architecture of the Conditional Neural Cellular Automata (c-NCA). The class-conditional vector $\mathbf{c}$ is injected into the local perception loop, guiding the update dynamics.}
    \label{fig:architecture}
\end{figure}
\subsection{Stochastic Update Policy}
The core dynamics are governed by a neural network $\Phi$ (parametrized as two $1 \times 1$ convolutional layers) that maps the perceived environment to a state update $\Delta S$. To simulate the asynchronous nature of biological systems and prevent oscillatory patterns, we apply a stochastic update mask $M_{stoch} \sim \text{Bernoulli}(p=0.5)$. The temporal evolution is defined as:
\begin{align}
    \Delta S &= \Phi(\mathbf{x}_{in}; \theta) \\
    S_{t+1} &= S_t + M_{stoch} \odot \Delta S
\end{align}
where $\odot$ denotes the Hadamard product. The network $\Phi$ utilizes a ReLU activation for the hidden layer and is initialized such that the final output weights are zero, ensuring $S_{t+1} \approx S_t$ at initialization (Identity mapping).

\subsection{Living Mask and Loss Function}
To constrain the growth and prevent background noise accumulation, we enforce an "Alive Masking" protocol. A cell is considered "alive" if its alpha value $\alpha > 0.1$ or if it has a mature neighbor. Cells failing this criterion have their states reset to zero.

The optimization objective is the Mean Squared Error (MSE) between the visible channels of the final state $S_T$ (at $T=64$) and the target RGBA image $Y$:
\begin{equation}
    \mathcal{L}(\theta) = \frac{1}{N} \sum_{i=1}^{N} || S_{T, visible}^{(i)} - Y^{(i)} ||_2^2
\end{equation}
This loss allows gradients to backpropagate through time (BPTT), adjusting the local rules to satisfy the global structural constraints.

\section{Experimental Setup}

To validate the efficacy of the proposed Conditional NCA, we conducted a series of experiments focusing on the model's ability to reliably grow target digits from a generic seed. In this section, we detail the dataset preparation, network architecture, and training protocols used.

\subsection{Dataset and Preprocessing}
We utilized the MNIST handwritten digit dataset, consisting of 60,000 training images and 10,000 test images. Since NCA operates in a continuous RGBA space, the standard grayscale images ($28 \times 28 \times 1$) were transformed into 4-channel target states ($28 \times 28 \times 4$). 
The transformation is defined as follows: the RGB channels are replicated from the grayscale intensity, and the Alpha channel ($\alpha$) is binarized such that $\alpha=1.0$ for foreground pixels (digit) and $\alpha=0.0$ for the background. This explicitly tasks the model with learning both the visual texture and the structural silhouette of the digit.

\subsection{Model Architecture and Hyperparameters}
The Update Network $\Phi$ is implemented as a fully convolutional network with a receptive field of $3 \times 3$. We employed a hidden dimension of 128 units. To ensure stability at the start of training, the weights of the final output layer were initialized to zero, ensuring that the initial updates $\Delta S$ are null, effectively starting the growth process from a stable identity state. Table \ref{tab:hyperparams} summarizes the core hyperparameters used in our experiments.

\begin{table}[h]
\centering
\caption{Hyperparameter Configuration}
\label{tab:hyperparams}
\begin{tabular}{lc}
\toprule
\textbf{Parameter} & \textbf{Value} \\
\midrule
Grid Size ($H \times W$) & $28 \times 28$ \\
State Channels ($C$) & 16 \\
Hidden Channels & 128 \\
Growth Steps ($T$) & 64 \\
Batch Size & 64 \\
Learning Rate & $1 \times 10^{-3}$ \\
Optimizer & Adam \\
Training Epochs & 25 \\
Stochastic Rate ($p$) & 0.5 \\
\bottomrule
\end{tabular}
\end{table}

\subsection{Training Protocol}
The model was implemented using the PyTorch framework and trained on a single NVIDIA GPU. Unlike standard NCA training which often employs a "sample pool" (replay buffer) to maintain persistent states, we adopted a \textit{de novo} growth strategy. In each iteration, a batch of blank seeds is initialized with a single activation at the center $(14, 14)$. The model is then run for $T=64$ steps. This rigorous protocol forces the model to learn the complete morphogenetic trajectory from scratch at every pass, ensuring that the learned rules are robust and deterministic given the initial condition.

The gradients were clipped at a norm of 1.0 to prevent exploding gradients, a common instability in recurrent dynamical systems.

\section{Results and Analysis}

In this section, we evaluate the dynamic morphogenesis of the trained c-NCA model. We move beyond qualitative observation to provide rigorous quantitative assessments of generation quality, stability, and the necessity of stochastic dynamics.

\subsection{Full Morphogenetic Trajectory}
\begin{figure}[H] 
    \centering
    \includegraphics[width=0.9\linewidth]{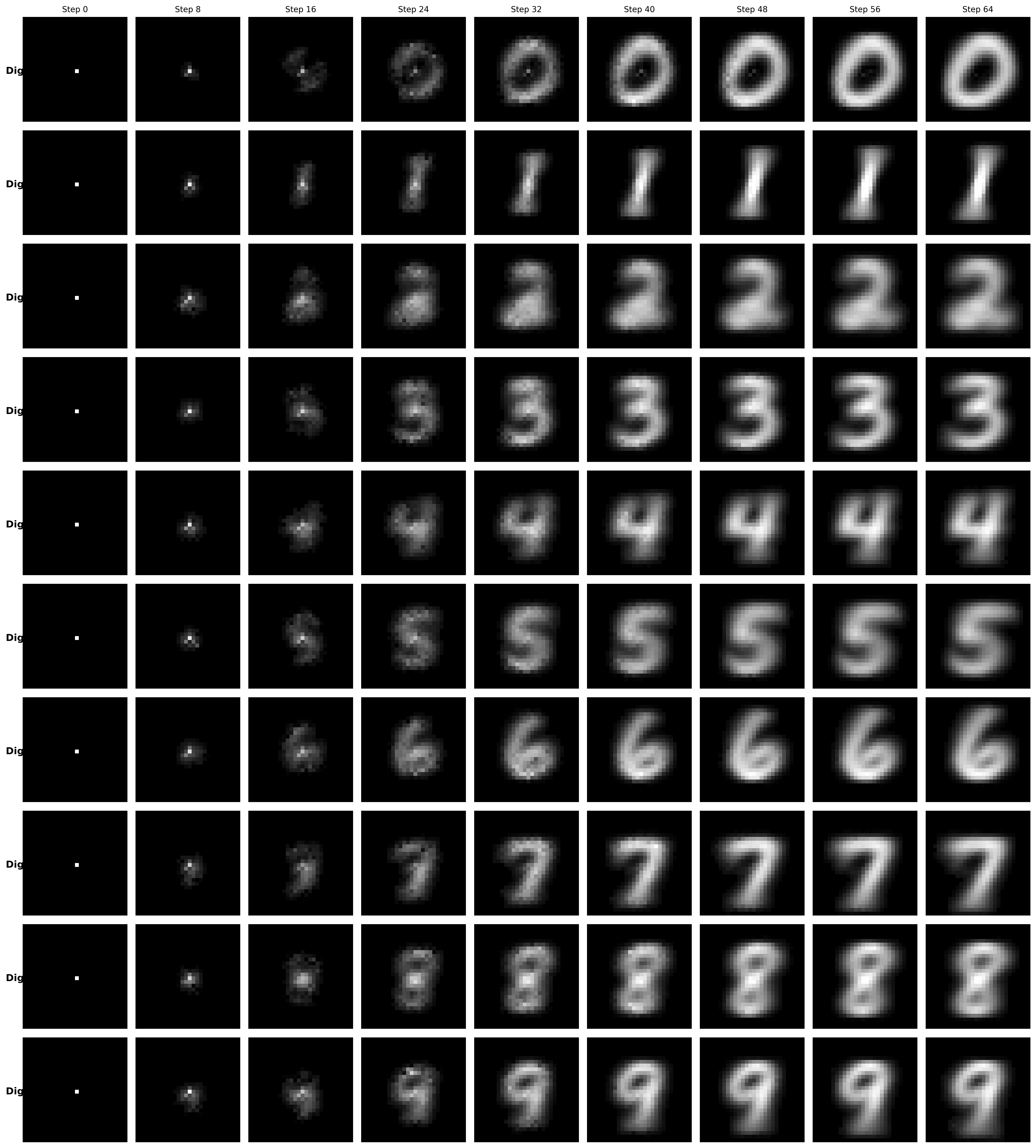} 
    \caption{Complete morphogenetic evolution of MNIST digits. All digits originate from the same generic seed. The condition vector guides the cellular automata to diverge into distinct topological attractors by Step 64.}
    \label{fig:full_growth}
\end{figure}

Figure \ref{fig:full_growth} provides a comprehensive visualization of the growth dynamics for all ten digit classes. Each row corresponds to a specific target class, conditioned by its one-hot vector $\mathbf{c}$. The visualization reveals a unified developmental timeline:
\begin{itemize}
    \item \textbf{Common Ancestry ($t=0 \to 8$):} All digits start as a single active pixel.
    \item \textbf{Symmetry Breaking ($t=8 \to 24$):} Driven by the condition vector, the circular blob deforms to match the topological requirements of the target digit.
    \item \textbf{Refinement ($t=32 \to 64$):} The boundaries sharpen, and the system settles into a stable attractor.
\end{itemize}

\subsection{Robustness to Stochastic Degradation}
\begin{figure}[H]
    \centering
    \includegraphics[width=0.55\linewidth]{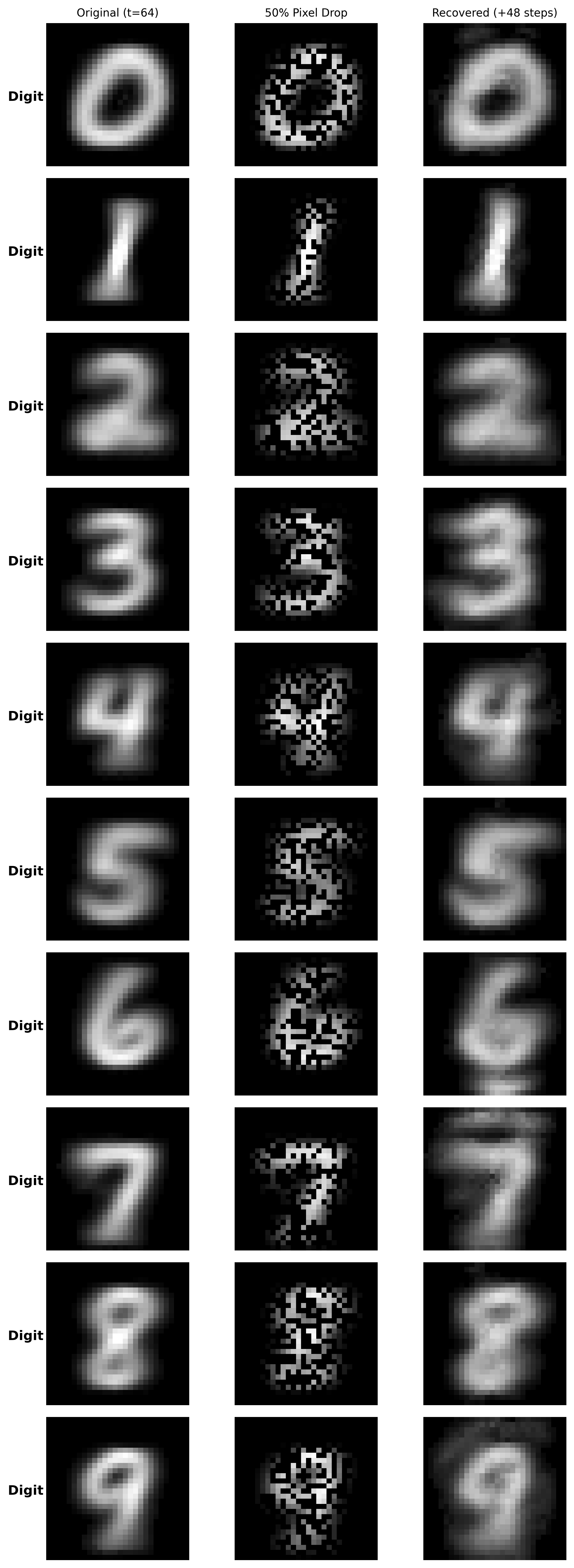} 
    \caption{Recovery from 50\% stochastic pixel dropout. The middle column depicts the state immediately after damage. The right column shows the successful reconstruction after 48 steps.}
    \label{fig:noise_repair}
\end{figure}

To evaluate resilience, we subjected the mature digits ($t=64$) to severe damage, instantaneously dropping 50\% of active pixels (Figure \ref{fig:noise_repair}). The system was allowed to evolve for 48 recovery steps. As evidenced, the recovery is nearly perfect, suggesting the c-NCA functions as a robust \textit{auto-associative memory}.

\subsection{Quantitative Evaluation}
To rigorously assess the generative quality, we employed two complementary metrics (summarized in Table \ref{tab:metrics}):

\textbf{1. Discriminative Accuracy:} We trained a standard LeNet-5 classifier on the original MNIST dataset to act as an external judge. When evaluated on 1,000 digits generated by our c-NCA from random seeds, the classifier achieved a \textbf{Recognition Accuracy of 96.30\%}. This confirms that the local update rules converge to topologically valid and distinguishable digit attractors.

\textbf{2. Structural Similarity (SSIM):} The average SSIM score between generated digits and random test samples was \textbf{0.4826}. While lower than reconstruction tasks, this is expected for generative models, as the c-NCA learns a canonical "ideal" shape rather than memorizing specific handwritten variations.

\begin{table}[H]
\centering
\caption{Quantitative Performance Metrics of c-NCA on MNIST}
\label{tab:metrics}
\begin{tabular}{lcl}
\toprule
\textbf{Metric} & \textbf{Value} & \textbf{Interpretation} \\
\midrule
Generation Accuracy & 96.30\% & Validated by LeNet-5 Classifier \\
Mean Confidence & 94.76\% & High semantic clarity of digits \\
Stability (MSE) & 0.0297 & Long-term structural persistence \\
SSIM & 0.4826 & Generative diversity vs. Test set \\
\bottomrule
\end{tabular}
\end{table}

\vspace{0.3em}
\noindent \textbf{3. Model Efficiency:} A striking advantage of our approach is its parameter efficiency. The entire c-NCA model consists of only \textbf{10,048 trainable parameters}. In stark contrast, traditional generative models like DCGAN typically require millions of parameters (e.g., $\approx 3 \times 10^6$) to model similar distributions. This makes the c-NCA orders of magnitude more lightweight, highlighting the efficiency of defining complexity through local rules rather than global weights.

\begin{figure}[H]
    \centering
    \includegraphics[width=0.45\linewidth]{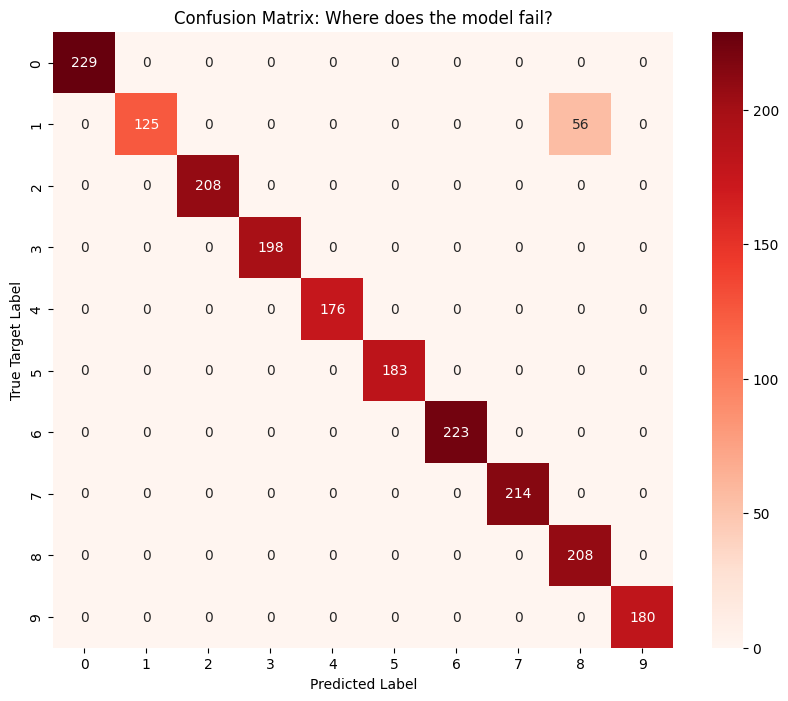} 
    \caption{Confusion Matrix of the c-NCA model. The strong diagonal indicates high accuracy. The specific off-diagonal cluster (Row '1', Column '8') highlights the structural over-growth artifact discussed in the error analysis.}
    \label{fig:confusion_matrix}
\end{figure}

\vspace{0.3em}
\noindent \textbf{4. Error Analysis:} While the overall recognition accuracy is high, the confusion matrix (see Figure \ref{fig:confusion_matrix}) reveals a specific, systematic failure mode. The dominant error occurs where the target digit '1' is misclassified as '8' (approx. 30\% of '1' samples). This suggests a phenomenon of \textit{structural over-growth}: since the '1' relies on inhibiting horizontal expansion, stochastic fluctuations sometimes allow lateral artifacts to form, which the classifier misinterprets as the closed loops of an '8'. Other classes exhibit minimal confusion, confirming the stability of more complex topologies.

\subsection{Stability and Semantic Confidence}
Beyond recognition, we evaluated the "homeostasis" of the generated structures:
\begin{itemize}
    \item \textbf{Semantic Confidence:} The LeNet-5 judge assigned a mean softmax probability of \textbf{94.76\%} to the generated digits, indicating that the structures contain strong, unambiguous features.
    \item \textbf{Homeostatic Stability:} We continued the simulation up to $t=128$. The Mean Squared Error (MSE) between $t=64$ and $t=128$ was \textbf{0.0297}. This low residual confirms the system reaches a stable \textit{dynamic equilibrium}, "breathing" slightly due to stochasticity but maintaining global topology.
\end{itemize}

\subsection{Ablation Study: The Role of Stochasticity}
To investigate the necessity of the asynchronous update rule, we performed an inference-time ablation by setting the stochastic update probability to $p=1.0$ (deterministic), compared to the standard $p=0.5$.

\begin{figure}[H]
    \centering
    \includegraphics[width=0.85\linewidth]{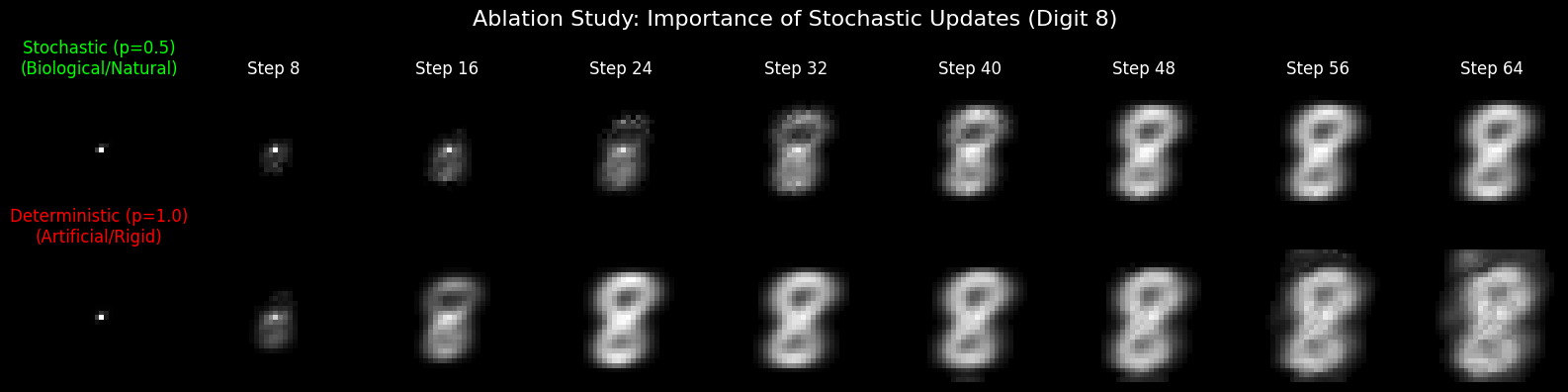} 
    \caption{Ablation study on the update rule. \textbf{Top:} Standard stochastic update ($p=0.5$) results in sharp topology. \textbf{Bottom:} Deterministic update ($p=1.0$) leads to blurred edges and synchronization artifacts.}
    \label{fig:ablation}
\end{figure}

As illustrated in Figure \ref{fig:ablation}, the removal of stochasticity resulted in a noticeable degradation. While the stochastic control (top row) produced sharp boundaries, the deterministic variant (bottom row) exhibited \textbf{"ghosting" artifacts and blurred edges}. This confirms that stochasticity acts as a crucial regularizer, preventing synchronization artifacts.

\clearpage
\section{Conclusion}

In this paper, we presented a novel framework for \textit{Conditional Structural Morphogenesis} using Neural Cellular Automata. By integrating a global class-conditional vector into the local perception-update loop, we demonstrated that a single, decentralized system can effectively encode and express multiple distinct topological attractors.

Our empirical results validate three key hypotheses:
\begin{enumerate}
    \item \textbf{Conditional Plasticity:} A lightweight network is capable of "switching" its developmental trajectory based entirely on an external signal.
    \item \textbf{Homeostatic Stability:} The generated structures exhibit stability over time, maintaining their coherent topology even after the growth phase concludes.
    \item \textbf{Intrinsic Resilience:} As demonstrated in the stochastic degradation experiments, the model possesses emergent self-repair capabilities, successfully reconstructing digits even after 50\% information loss without explicit training on damaged data.
\end{enumerate}

This work serves as a proof-of-concept that Generative AI can move beyond "constructing" images to "growing" them. Future work will explore scaling this architecture to richer RGB datasets such as CIFAR-10.

\bibliographystyle{unsrt}
\bibliography{references}
\end{document}